\documentclass[letterpaper]{article} 
\usepackage[preprint]{aaai2027}  
\usepackage[hyphens]{url}  
\usepackage{graphicx} 
\usepackage{natbib}  
\usepackage{caption} 
\usepackage{algorithm}
\usepackage{algorithmic}
\usepackage{booktabs}   
\usepackage{colortbl}   
\usepackage{xcolor}     

\usepackage{float}

\usepackage{newfloat}
\usepackage{listings}
\DeclareCaptionStyle{ruled}{labelfont=normalfont,labelsep=colon,strut=off} 
\floatstyle{ruled}
\newfloat{listing}{tb}{lst}{}
\floatname{listing}{Listing}

\definecolor{groupgray}{RGB}{241, 241, 241}
\definecolor{methodgray}{RGB}{248, 248, 248}

\newcolumntype{Y}{>{\columncolor{succyellow}}c}
\usepackage{booktabs}

\usepackage{amsmath}
\usepackage{amssymb}

\title{Bidirectional Context Self-Distillation for Reinforcement Learning of Skill-Based LLM Agents}
\author{
    \large Tianjun Pan\textsuperscript{\rm 1}\equalcontrib, Yuan Li\textsuperscript{\rm 2}\equalcontrib, Hongda Wang\textsuperscript{\rm 4}, Linbo Jin\textsuperscript{\rm 4}\corresponding, 
    \large Mengfei Song\textsuperscript{\rm 4}, Lei Gao\textsuperscript{\rm 1},\\ Qiming Shi\textsuperscript{\rm 2}, Shaokang Fu\textsuperscript{\rm 4}, 
    \large Jiarong Zhao\textsuperscript{\rm 3}, Chengyu Wang\textsuperscript{\rm 4}, Chengfu Huo\textsuperscript{\rm 4}
}
\affiliations{
    \textsuperscript{\rm 1} Fudan University  \textsuperscript{\rm 2} Zhejiang University  \\ \textsuperscript{\rm 3} East China Normal University  \textsuperscript{\rm 4} Alibaba Group
}

\begin{document}

\maketitle

\begin{abstract}

External natural-language skills provide large language model (LLM) agents with reusable and editable guidance for solving complex tasks. Yet their effectiveness depends not only on skill quality, but also on whether the policy can translate the provided guidance into appropriate actions. However, methods specifically designed to improve this skill-utilization ability remain largely underexplored. 
In practice, skill-based agents are commonly trained with reinforcement learning objectives centered on task-level rewards, which offer limited supervision and struggle to capture subtle differences in how effectively the policy uses the provided skills. 
We propose \textbf{BCSD} (\textbf{B}idirectional \textbf{C}ontext \textbf{S}elf-\textbf{D}istillation), a framework that combines self-distillation with reinforcement learning to train LLM agents to use external skills more effectively. Unlike prior self-distillation methods that rely on a single privileged context, BCSD evaluates each trajectory from two complementary skill-context views. The augmented view introduces higher-level Meta-Skill guidance, while the reduced view prunes general guidance to highlight task-specific skills. Their complementary token-level signals are combined to rescale the RL advantage.
Experiments on ALFWorld and WebShop demonstrate that BCSD achieves the strongest overall performance across model scales, enabling agents to utilize external skills more effectively. Ablation studies further verify the complementary contributions of the augmented and reduced context views. Code will be released to ensure full reproducibility.

\end{abstract}


\section{Introduction}





Large language model (LLM) agents \cite{react,reflexion} increasingly use natural-language skills as reusable knowledge across related tasks \cite{agent_skill_survey,sok}. By conditioning on relevant skills, agents can reuse prior experience instead of solving each task from scratch \cite{skillrl,skill1,skillclaw}.

\begin{figure}[t]
\centering
\includegraphics[width=0.9\columnwidth]{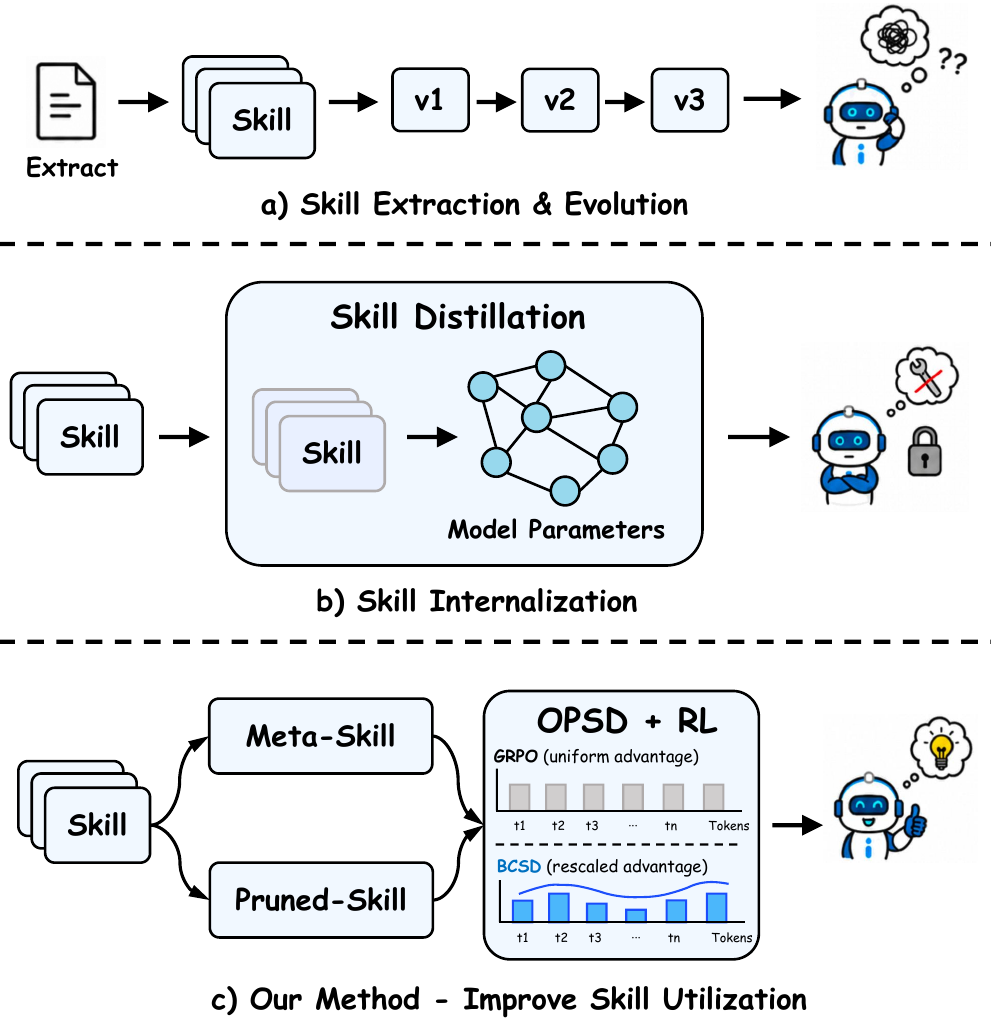}
\caption{Comparison of three paradigms for skill-based agents. Unlike skill evolution or internalization, our method improves the model’s ability to utilize explicit external skills.}
\label{fig1}
\end{figure}

Recent work on skill-augmented LLM agents has followed two main directions. One focuses on generating and refining external skills to improve their quality \cite{skillclaw,skillevolver,trace2skill}, but pays limited attention to whether agents can reliably use them in long-horizon tasks. Recent evidence further suggests that providing high-quality skills does not necessarily lead to effective skill utilization \cite{skillsbench}. The other internalizes skills into model parameters, reducing reliance on external skills at the cost of their explicitness, editability, and transferability \cite{sdar,skillsd,opid}. These two directions are summarized in Figure~\ref{fig1}(a) and (b), respectively.

This motivates an important yet underexplored question: \textit{how can agents learn to better use skills provided explicitly as external context, rather than internalizing them into model parameters?}
Although prior work has applied reinforcement learning to skill-conditioned policies \cite{skillrl}, long-horizon agent tasks typically provide only sparse outcome rewards. Such coarse supervision cannot identify which decisions are genuinely informed by the provided skills, making it difficult for the policy to learn effective skill utilization.

To obtain more fine-grained supervision, on-policy self-distillation (OPSD) has recently emerged as a promising training paradigm \cite{opsd,rlsd,sdar}. However, the token-level signal produced by a privileged self-teacher is not always reliable. Recent studies report information leakage and unstable optimization \cite{rlsd,sdar}, and show that the resulting teacher--student gap may emphasize context-induced stylistic shifts or shortcut tokens rather than task-relevant ones \cite{rlcsd,antiselfdistillation}. These findings cast doubt on whether a single context-induced probability gap can provide reliable credit for each token in long-horizon agent tasks.

These observations motivate us to move beyond a single contextual teacher. Inspired by MOPD's integration of multiple teacher signals \citep{mopd} and the anchoring role of reference policies in reinforcement learning \citep{instructgpt,deepseekmath}, we translate the integration of multiple teachers and the use of a shared reference from parameter space to context space. Multiple contextual teachers provide complementary estimates of token-level guidance, while using the same policy under different contexts makes their signals directly comparable.

We propose \textbf{BCSD} (\textbf{B}idirectional \textbf{C}ontext \textbf{S}elf-\textbf{D}istillation), which combines self-distillation with reinforcement learning for skill-based agents. Unlike existing approaches that construct an on-policy self-teacher only by adding privileged context, BCSD distills from two complementary views. The augmented view adds high-level Meta-Skill guidance on how the provided skills should be applied, while the pruned view condenses the surrounding context but preserves the complete task-specific skill, providing a focused reference for whether each decision follows it. The same policy acts as a self-teacher under both views and produces token-level signals relative to the original skill context. BCSD combines these signals through a weighted aggregation and uses them to rescale the GRPO advantage without reversing its optimization direction. In this way, the two views operate jointly and provide more reliable supervision for the agent to learn to use the provided skills.

We evaluate BCSD on two long-horizon agent benchmarks, ALFWorld \cite{alfworld} and WebShop \cite{webshop}, across multiple model scales. Experiments show that BCSD consistently outperforms representative baselines. Further analyses confirm that combining the two contextual views is both necessary and effective.
These results demonstrate the effectiveness of BCSD for training skill-based agents. Our contributions are summarized as follows:

\begin{itemize}
\item We identify improving LLM agents' use of external skills, rather than internalizing skill knowledge into model parameters, as an important direction for skill-based agent.

\item We propose \textbf{BCSD}, a method that derives complementary token-level guidance from augmented and reduced skill contexts and integrates it into GRPO advantage to improve external skill utilization for LLM agents.

\item Experiments on ALFWorld and WebShop show that BCSD achieves the strongest overall performance across model scales, validating its effectiveness for skill-based LLM agents training. 
\end{itemize}

\section{Related Work}

\subsection{Skill-Augmented LLM Agents}

\begin{figure*}[t]
    \centering
    \includegraphics[width=\textwidth]{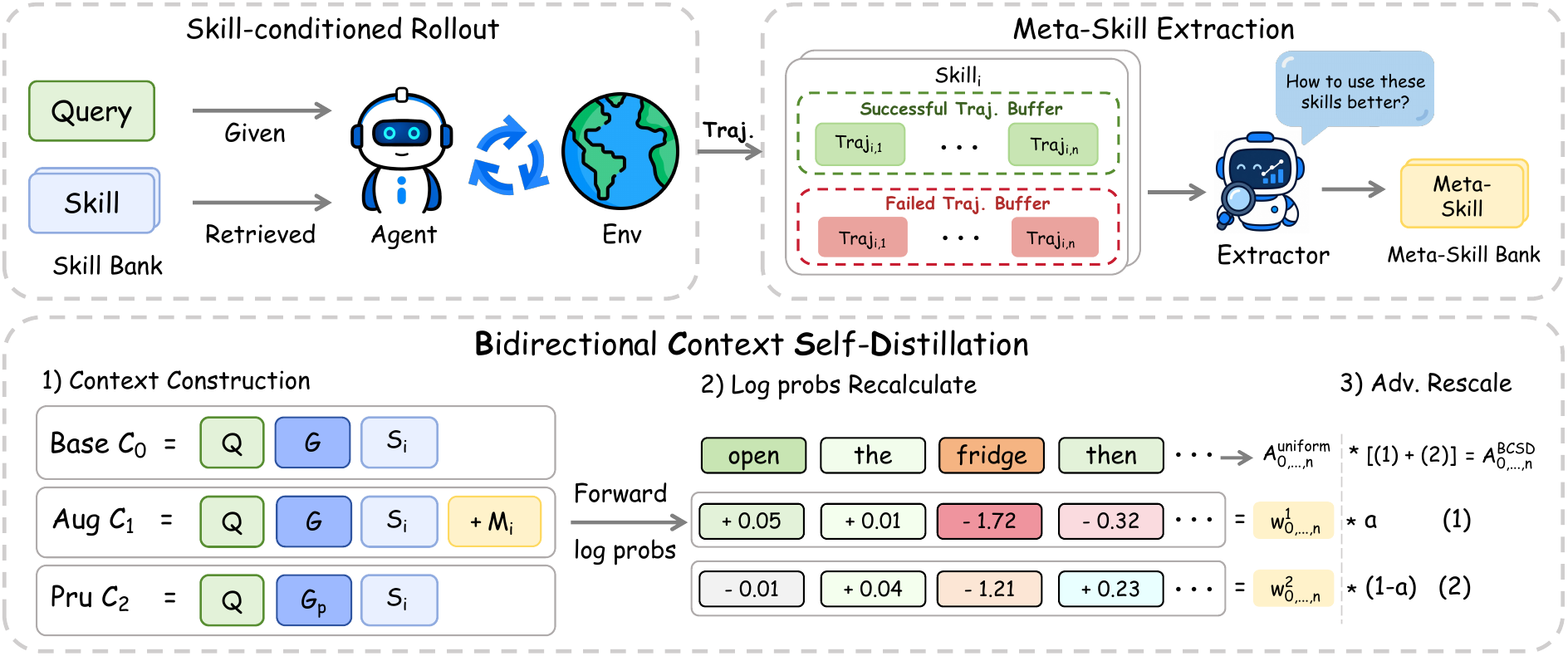}
    \caption{\textbf{Overview of BCSD.} It combines skill-conditioned rollouts, trajectory-based meta-skill extraction, and bidirectional context self-distillation for token-wise advantage rescaling.}
    \label{fig:figure2}
\end{figure*}

Natural-language skills equip LLM agents with reusable procedural knowledge, greatly improving efficiency. Recent work studies their construction, refinement, utilization, and transfer. Early studies maintain skills as external text that can be retrieved into the agent's context. SkillRL \citep{skillrl} constructs a hierarchical SkillBank and jointly evolves it with the policy through skill-augmented training. SkillEvolver \citep{skillevolver}, SkillAudit \citep{skillaudit}, and SkillHone \citep{skillhone} improve skill repositories through generation, auditing, and revision. However, providing diverse external knowledge does not ensure its effective use. Other methods internalize skills into model parameters. Skill-SD \citep{skillsd} distills a skill-conditioned teacher into the student policy. SDAR \citep{sdar} combines skill-conditioned self-distillation with reinforcement learning and adaptively gates the distillation signal. SIRI \citep{siri} extracts skills from policy trajectories and transfers skill-guided action tokens into the standard policy. While removing the need for explicit inference-time skills, internalization reduces their inspectability, editability, and transferability. In contrast, BCSD retains external skills while training the policy to utilize them more effectively. 

\subsection{On-Policy Self-Distillation Training Paradigm} 
On-policy distillation (OPD) applies token-level teacher supervision to student rollouts \citep{gkd}.
Recent studies extend it to agents through step-wise distillation \citep{sod} and temporal curricula \citep{tcod}. OPD has further been extended to on-policy self-distillation (OPSD), where the same model acts as teacher and student under different contexts \citep{opsd,sdpo}, with applications to reasoning compression \citep{crisp} and long-context modeling \citep{opsdl}. 
More recent work combines OPSD with RL to complement sparse rewards. RLSD \citep{rlsd} uses teacher--student discrepancies for token-level credit assignment, while Skill-SD \citep{skillsd} derives skill-conditioned supervision from the agent's trajectories. SDAR \citep{sdar} adaptively gates distillation signals, HERO \citep{hero} grounds them in environment observations, and StepOPSD \citep{stepopsd} assigns step-level credit. OPID \citep{opid} further introduces skill-aware routing. BCSD extends this paradigm to improve external-skill utilization during reinforcement learning while preserving skills as explicit inference-time context.

\section{Methods}
We argue that skill-based LLM agents should retain natural-language skills as explicit context during both training and inference. Based on this principle, we develop BCSD to improve the policy's ability to use external skills through bidirectional context self-distillation and reinforcement learning. Figure~\ref{fig:figure2} presents an overview of the framework.
\subsection{Problem Setup}

We formulate a skill-based long-horizon agentic task as a partially observable decision process \cite{rlida}. At interaction step $h$, the agent receives an observation $o_h$ and generates an action $a_h$ conditioned on the preceding interaction history. A trajectory $\tau=(o_1,a_1,\ldots,o_H,a_H)$ receives a terminal outcome reward $R(\tau)$ after task completion. For the Skill setup, we follow SkillRL \citep{skillrl} and adopt a hierarchical SkillBank setup. For each query $Q$ of task type $c$, the SkillBank provides a general skill $G$ and a retrieved task-specific skill $S_c$. The base context is defined as:
\begin{equation}
    C_0 = (Q, G, S_c).
\end{equation}
For each query, let $\tau_i$ denote the i-th trajectory in a group of K on-policy trajectories sampled by the behavior policy under the base context:
\begin{equation}
    \{\tau_i\}_{i=1}^{K}
    \sim \pi_{\theta}(\cdot \mid C_0),
    \qquad
    r_i = R(\tau_i),
    \label{eq:rollout}
\end{equation}
where $i$ indexes the sampled trajectories and $r_i$ is the outcome reward of $\tau_i$. Standard GRPO \cite{deepseek-r1} computes the group-relative advantage:
\begin{equation}
    A_i^{\mathrm{GRPO}}
    =
    \frac{r_i-\mu_r}{\sigma_r+\epsilon},
\end{equation}
where $\mu_r$ and $\sigma_r$ are the mean and standard deviation of the rewards within the group.

Let $y_i=(y_{i,1},\ldots,y_{i,T_i})$ denote the response-token sequence in $\tau_i$ and $y_{i,<t}$ its realized prefix. 
GRPO assigns the same trajectory-level advantage $A_i^{\mathrm{GRPO}}$ to every response token. BCSD instead evaluates the same sequence under two complementary context views, $C_1$ and $C_2$, and uses the resulting token-level signals to rescale the original advantage.

\subsection{Meta-Skill Extraction}

\paragraph{Definition.}
The external skills in $C_0$ provide procedural knowledge for solving a task, but they do not reflect how the current policy succeeds or fails in applying that knowledge. 
We define \textbf{Meta-Skill} as high-level textual guidance that summarizes effective and ineffective ways of applying skills across rollout trajectories. For each task-specific skill $S_c$, we maintain a corresponding Meta-Skill $M_c$.
Unlike the skills stored in the SkillBank, $M_c$ captures policy-dependent guidance on how $S_c$ should be used. It is provided only to the teacher as privileged context during training and is not included in the student context $C_0$ or used during inference.

\paragraph{Extraction and Refresh.}
Before training, we collect rollout trajectories from multiple queries associated with each skill $S_c$. We divide them into successful and failed evidence sets, denoted by $\mathcal{E}_{c,+}^{(0)}$ and $\mathcal{E}_{c,-}^{(0)}$, respectively. All evidence is collected from on-policy training rollouts; validation checkpoints only determine when the refresh is triggered and do not provide extraction data. A frozen external LLM contrasts the two sets and initializes the meta-skill as
\begin{equation}
    M_c^{(0)}
    =
    \operatorname{Extract}_{\mathrm{LLM}}
    \left(
        S_c,
        \mathcal{E}_{c,+}^{\mathrm{(0)}},\mathcal{E}_{c,-}^{\mathrm{(0)}}
    \right).
\end{equation}

To track the policy’s evolving skill-utilization patterns, we refresh each Meta-Skill after every validation window using the trajectories collected during window $k$:
\begin{equation}
    M_c^{(k+1)}
    =
    \operatorname{Extract}_{\mathrm{LLM}}
    \left(
        S_c,
        M_c^{(k)},
        \mathcal{E}_{c,+}^{(k)},
        \mathcal{E}_{c,-}^{(k)}
    \right).
    \label{eq:meta_refresh}
\end{equation}
After each refresh, the updated Meta-Skill is used for the next training window and the evidence buffers are reset.

\subsection{Bidirectional Context Self-Distillation}

Existing OPSD methods typically derive token-level supervision from a single privileged context, which may induce context-specific shifts in the teacher distribution and miscalibrate its guidance \citep{oglssd,edgeopd}. BCSD instead constructs two complementary views around the base context $C_0$.

\paragraph{Complementary Context Construction.}
The augmented context extends $C_0$ with the Meta-Skill $M_c$ associated with the current task type $c$:
\begin{equation}
    C_1 = (Q, G, S_c, M_c).
\end{equation}
The Meta-Skill provides high-level skill utilization guidance derived from the evolving behavior of the skill-based policy. 
In the opposite direction, we use an external LLM to remove redundant content from $G$ while retaining its essential procedural guidance:
\begin{equation}
    G_p = \operatorname{Prune}_{\mathrm{LLM}}(G).
\end{equation}
This pruning is performed once before training, after which $G_p$ remains fixed. The resulting pruned context is
\begin{equation}
    C_2 = (Q, G_p, S_c).
\end{equation}
By replacing $G$ with $G_p$ while preserving $S_c$, the pruned view reduces distracting information and offers a focused reference for task-specific skill usage \citep{opsdl}.
The augmentation from $C_0$ to $C_1$ and the pruning from $C_0$ to $C_2$ constitute the two directions of BCSD.

\paragraph{Gap-Weighted Advantage Rescaling.}
For each trajectory $\tau_i$ sampled under $C_0$, we keep the realized response sequence $y_i$ fixed and use the same policy to re-score its tokens under $C_1$ and $C_2$. This yields two token-level log-probability gaps:
\begin{align}
    \Delta_{i,t}^{\mathrm{aug}}
    &=
    \log \pi_\theta(y_{i,t}\mid C_1,y_{i,<t})
    -
    \log \pi_\theta(y_{i,t}\mid C_0,y_{i,<t}),
    \\
    \Delta_{i,t}^{\mathrm{pru}}
    &=
    \log \pi_\theta(y_{i,t}\mid C_2,y_{i,<t})
    -
    \log \pi_\theta(y_{i,t}\mid C_0,y_{i,<t}).
\end{align}
The two gaps quantify how adding high-level guidance or removing redundant information changes the policy's support for each realized token. We combine them as
\begin{equation}
    \Delta_{i,t}
    =
    \alpha\Delta_{i,t}^{\mathrm{aug}}
    +(1-\alpha)\Delta_{i,t}^{\mathrm{pru}},
    \label{eq:bcsd_gap}
\end{equation}
where $\alpha$ determines the relative weights of the augmented and pruned views.

Unlike prior credit-assignment methods that explicitly estimate turn- or segment-level advantages \citep{gigpo,sapo,turnppo,salt}, 
BCSD incorporates self-distillation through token-wise advantage rescaling rather than a separate loss, following the design of prior work \citep{rlsd}.
We design and use the bidirectional gap $\Delta_{i,t}$ to modulate the token-level advantage without replacing it:
\begin{equation}
    w_{i,t}
    =
    \operatorname{clip}
    \left(
        \exp\left[
            \operatorname{sign}(A_i^{\mathrm{GRPO}})
            \Delta_{i,t}
        \right],
        1-c,1+c
    \right),
    \label{eq:bcsd_weight}
\end{equation}
\begin{equation}
    A^{\mathrm{BCSD}}_{i,t}
    =
    A_i^{\mathrm{GRPO}}
    \left[
    (1-\lambda_n)+\lambda_n w_{i,t}
    \right].
    \label{eq:bcsd_advantage}
\end{equation}
BCSD changes only the magnitude of the advantage while preserving the update direction determined by the trajectory-level reward. The coefficient $\lambda_n\in[0,1]$ controls the rescaling strength at training step $n$ and is linearly decayed from $\lambda_0$ to zero, providing strong contextual guidance early in training while gradually diminishing.

\subsection{Policy Optimization Objective}

BCSD follows the standard clipped GRPO objective and replaces the original trajectory-level advantage with $A_{i,t}^{\mathrm{BCSD}}$. The token-level importance ratio is
\begin{equation}
    \rho_{i,t}(\theta)
    =
    \frac{
        \pi_\theta(y_{i,t}\mid C_0,y_{i,<t})
    }{
        \pi_{\theta_{\mathrm{old}}}(y_{i,t}\mid C_0,y_{i,<t})
    }.
\end{equation}
The final policy loss is
\begin{equation}
\begin{aligned}
    \mathcal{L}_{\mathrm{BCSD}}(\theta)
    =&-\mathbb{E}_{i,t}\Big[
    \min\Big(
        \rho_{i,t}(\theta)A_{i,t}^{\mathrm{BCSD}},\\
    &\operatorname{clip}\big(
        \rho_{i,t}(\theta),1-\epsilon,1+\epsilon
    \big)A_{i,t}^{\mathrm{BCSD}}
    \Big)\Big]
    +\beta\mathcal{L}_{\mathrm{KL}},
\end{aligned}
\label{eq:bcsd_objective}
\end{equation}
where $\epsilon$ is the clipping coefficient, $\beta$ controls the strength of KL regularization, and $\mathcal{L}_{\mathrm{KL}}$ denotes the standard KL loss. The complete training procedure of BCSD with GRPO is summarized in Algorithm~\ref{alg:bcsd}.

\begin{algorithm}[t]
\caption{BCSD Algorithm}
\label{alg:bcsd}
\small
\begin{algorithmic}[1]
\REQUIRE Policy $\pi_\theta$, dataset $\mathcal{D}$, SkillBank
$\{G,\{S_c\}\}$, reward $R$, group size $K$, coefficients
$\alpha,\{\lambda_n\}$, refresh interval $N$
\ENSURE Trained policy $\pi_{\theta^*}$

\STATE Construct $G_p$ and initialize $\{M_c^{(0)}\}$ from
preloaded trajectories using a frozen external LLM
\STATE $k\leftarrow0$; initialize
$\mathcal{E}_{c,+}^{(0)},\mathcal{E}_{c,-}^{(0)}
\leftarrow\varnothing$ for all $c$

\FOR{optimization step $n=1,2,\ldots$}
    \STATE Sample $\mathcal{B}\subset\mathcal{D}$ and set
    $\theta_{\mathrm{old}}\leftarrow\theta$
    \FOR{each $(Q,c)\in\mathcal{B}$}
        \STATE $C_0\leftarrow(Q,G,S_c)$,
        $C_1\leftarrow(Q,G,S_c,M_c^{(k)})$,
        $C_2\leftarrow(Q,G_p,S_c)$
        \STATE Sample
        $\{\tau_i\}_{i=1}^{K}
        \sim\pi_{\theta_{\mathrm{old}}}(\cdot\mid C_0)$
        and compute $\{r_i,A_i^{\mathrm{GRPO}}\}_{i=1}^{K}$
        \STATE Re-score the realized tokens under $C_0,C_1,C_2$
        using $\pi_{\theta}$
        \STATE Compute $\Delta_{i,t}$, $w_{i,t}$, and
        $A_{i,t}^{\mathrm{BCSD}}$ using
        Eqs.~\eqref{eq:bcsd_gap}--\eqref{eq:bcsd_advantage}
        \STATE Store skill-stripped trajectories in
        $\mathcal{E}_{c,+}^{(k)}$ or $\mathcal{E}_{c,-}^{(k)}$
    \ENDFOR

    \STATE Update $\theta$ by minimizing
    $\mathcal{L}_{\mathrm{BCSD}}$ in
    Eq.~\eqref{eq:bcsd_objective}

    \IF{$n\bmod N=0$}
        \STATE Refresh $\{M_c^{(k+1)}\}$ using
        Eq.~\eqref{eq:meta_refresh} and reset the evidence buffers
        \STATE $k\leftarrow k+1$
    \ENDIF
\ENDFOR
\RETURN $\pi_{\theta^*}$
\end{algorithmic}
\end{algorithm}

\begin{table*}[t]
\centering
\small
\setlength{\tabcolsep}{7.5pt}
\renewcommand{\arraystretch}{1.00}
\setlength{\aboverulesep}{0.25ex}
\setlength{\belowrulesep}{0.25ex}
\setlength{\cmidrulesep}{0.15ex}

\begin{tabular}{lccccccccc}
    \toprule
    & \multicolumn{7}{c}{\textbf{\textsc{ALFWorld}}}
    & \multicolumn{2}{c}{\textbf{\textsc{WebShop}}} \\
    \cmidrule(lr){2-8}
    \cmidrule(lr){9-10}
    \textbf{Method}
    & \textbf{Pick}
    & \textbf{Look}
    & \textbf{Clean}
    & \textbf{Heat}
    & \textbf{Cool}
    & \textbf{Pick2}
    & \textbf{All}
    & \textbf{Score}
    & \textbf{Succ.} \\
    \midrule

    \rowcolor{groupgray}
    \multicolumn{10}{>{\columncolor{groupgray}}l}
    {\textit{Qwen2.5-7B-Instruct}} \\

    Vanilla
    & 27.8 & 22.2 & 18.8 & 7.7 & 14.3 & 4.2
    & 17.2 & 4.2 & 0.0 \\

    Skill\_Prompt
    & 31.0 & 50.0 & 25.8 & 10.5 & 0.0 & 0.0
    & 17.2 & 0.7 & 0.0 \\

    Skill\_GRPO
    & 86.2 & \textbf{72.7} & \textbf{100.0} & \underline{90.9}
    & 65.5 & 65.4 & 78.9 & 74.1 & 64.1 \\

    Skill\_GRPO$^*$
    & \underline{90.9} & \underline{70.0} & 84.0 & 83.3
    & 69.6 & \textbf{76.0} & \underline{80.5} & 85.6 & 75.0 \\

    RLSD
    & 89.7 & 66.7 & 87.0 & 46.7 & 66.7 & 42.9
    & 68.0 & 84.2 & 70.3 \\

    Skill-SD
    & 72.4 & 16.7 & \underline{95.7} & 60.0 & 51.9 & 53.6
    & 64.1 & \textbf{86.1} & \underline{75.8} \\

    SDAR
    & \textbf{93.1} & 33.3 & \underline{95.7} & 80.0
    & \textbf{77.8} & 67.9 & \underline{80.5} & 85.0 & 73.4 \\

    \rowcolor{methodgray}
    \textbf{\textsc{BCSD}}
    & 89.7 & 66.7 & \textbf{100.0} & \textbf{93.3}
    & \underline{74.1} & \underline{71.4} & \textbf{83.6}
    & \underline{85.9} & \textbf{78.1} \\

    \midrule
    \rowcolor{groupgray}
    \multicolumn{10}{>{\columncolor{groupgray}}l}
    {\textit{Qwen2.5-3B-Instruct}} \\

    Vanilla
    & 58.3 & 22.2 & 9.4 & 7.7 & 7.1 & 12.5
    & 24.2 & 9.1 & 1.6 \\
    
    Skill\_Prompt
    & 44.8 & \underline{50.0} & 22.6 & 0.0 & 4.3 & 5.0
    & 19.5 & 2.5 & 0.0 \\

    Skill\_GRPO
    & \textbf{97.1} & \textbf{53.8} & 54.2 & 25.0 & 41.2
    & \underline{60.9} & 61.7 & 54.3 & 37.5 \\

    Skill\_GRPO$^*$
    & 90.9 & 42.9 & \underline{96.3} & \underline{69.2}
    & 76.9 & 45.5 & \underline{76.6} & 63.4 & 48.4 \\

    RLSD
    & 89.7 & 33.3 & 82.6 & 66.7 & 66.7 & 35.7
    & 66.4 & 77.4 & 53.1 \\

    Skill-SD
    & 82.8 & \underline{50.0} & 87.0 & 60.0
    & \textbf{81.5} & 57.1 & 73.4
    & \underline{77.9} & 53.9 \\

    SDAR
    & 82.8 & 33.3 & 78.3 & \textbf{80.0} & 74.1 & 53.6
    & 71.1 & \underline{77.9} & \underline{57.0} \\

    \rowcolor{methodgray}
    \textbf{\textsc{BCSD}}
    & \underline{93.1} & \underline{50.0} & \textbf{100.0}
    & 66.7 & \underline{77.8} & \textbf{75.0}
    & \textbf{82.0} & \textbf{82.3} & \textbf{66.4} \\

    \midrule
    \rowcolor{groupgray}
    \multicolumn{10}{>{\columncolor{groupgray}}l}
    {\textit{Qwen3-1.7B}} \\
    
    Vanilla
    & 47.2 & 22.2 & 0.0 & 0.0 & 21.4 & \underline{29.2}
    & 22.7 & 40.1 & 3.9 \\
    
    Skill\_Prompt
    & \textbf{55.2} & 33.3 & 58.1 & 21.1
    & 13.0 & 20.0 & 36.7 & 25.3 & 2.3 \\
    
    Skill\_GRPO
    & 38.1 & \textbf{55.6} & 23.3 & 21.4 & 23.3
    & \underline{29.2} & 28.9 & 61.0 & 26.6 \\
    
    Skill\_GRPO$^*$
    & 42.9 & 27.3 & 51.9 & 16.7 & \textbf{61.5}
    & \textbf{43.3} & \underline{43.0}
    & \underline{75.1} & 48.4 \\
    
    RLSD
    & \underline{51.7} & 33.3 & \underline{65.2} & 33.3
    & 25.9 & 7.1 & 35.9 & 69.0 & \textbf{51.6} \\
    
    Skill-SD
    & \textbf{55.2} & 33.3 & 52.2 & 40.0
    & \underline{44.4} & 3.6 & 38.3 & 72.9 & 46.9 \\
    
    SDAR
    & 40.0 & \underline{50.0} & 53.8 & \textbf{57.1}
    & 35.0 & 26.3 & 42.2 & 41.6 & 28.1 \\

    \rowcolor{methodgray}
    \textbf{\textsc{BCSD}}
    & \underline{51.7} & 33.3 & \textbf{69.6}
    & \underline{53.3} & \underline{44.4} & 25.0
    & \textbf{46.9} & \textbf{76.8} & \underline{50.8} \\

    \bottomrule
\end{tabular}

\caption{
    Main results on \textsc{ALFWorld} and \textsc{WebShop}.
    \textsc{ALFWorld} reports subtask-specific and overall success rates (\%), while
    \textsc{WebShop} reports task score and success rate (\%).
    The best and second-best results within each model group are highlighted in
    \textbf{bold} and \underline{underline}, respectively. $^*$ denotes evaluation with skills. All results are obtained from our own experiments.
}
\label{tab:main_results}

\end{table*}

\section{Experiments}
\subsection{Experimental Setup}

\paragraph{Benchmarks.}
We evaluate BCSD on two long-horizon interactive agent benchmarks: ALFWorld \citep{alfworld}, which requires agents to complete household tasks through textual interactions, and WebShop \citep{webshop}, which evaluates product search and purchasing under user-specified constraints. We report the average success rate on ALFWorld and both task score and success rate on WebShop.

\paragraph{Baselines.}
We compare BCSD with prompting-only, outcome-based RL, and self-distillation baselines: Vanilla, Skill\_Prompt, Skill\_GRPO, Skill-SD \citep{skillsd}, RLSD \citep{rlsd}, and SDAR \citep{sdar}. All results are reproduced under the same hierarchical SkillBank and experimental setup to ensure fair comparison, with base skills retained during training and evaluation. For self-distillation baselines, the Meta-Skill serves as privileged context, while the student is conditioned only on the base skills. Skill\_GRPO$^*$ retains external skills at inference, whereas Skill\_GRPO evaluates the same policy without them.

\paragraph{Implementation details.}
We conduct experiments with Qwen2.5-3B-Instruct, Qwen2.5-7B-Instruct \cite{qwen25}, and Qwen3-1.7B \cite{qwen3}. Each model is trained for 150 steps with a rollout group size of $K=8$ and evaluated every 15 steps. For BCSD, we set $\alpha=0.9$ and linearly decay $\lambda_n$ from $0.1$ to zero. Meta-Skills are refreshed at each evaluation interval. All other settings follow RLSD \citep{rlsd}.

\subsection{Main Results}

\begin{table}[t]
\centering
\small
\setlength{\tabcolsep}{6pt}
\renewcommand{\arraystretch}{1.1}

\begin{tabular}{lcc}
    \toprule
    \textbf{Variant}
    & \textbf{\textsc{ALFWorld}}
    & \textbf{\textsc{WebShop}} \\
    \midrule

    \textbf{BCSD}
    & \textbf{82.0}
    & \textbf{66.4} \\

    \midrule
    \rowcolor{groupgray}
    \multicolumn{3}{l}{\textit{Teacher context view}} \\

    \quad w/o Meta-Skill teacher
    & 71.1
    & 50.0 \\

    \quad w/o Pruned-Skill teacher
    & 75.0
    & 57.0 \\

    \midrule
    \rowcolor{groupgray}
    \multicolumn{3}{l}{\textit{Rescaling coefficient schedule}} \\

    \quad $\lambda_n{=}\lambda_0$ (no decay)
    & 76.6
    & 52.3 \\

    \midrule
    \rowcolor{groupgray}
    \multicolumn{3}{l}{\textit{Meta-skill update}} \\

    \quad w/o Updating Meta-Skill
    & 78.1
    & 50.0 \\

    \bottomrule
\end{tabular}

\caption{
    Ablation results on \textsc{ALFWorld} and \textsc{WebShop}
    with the Qwen2.5-3B-Instruct.
    We report success rates (\%).
}
\label{tab:ablation}

\end{table}

Table~\ref{tab:main_results} presents the main results across different model scales. BCSD achieves the strongest aggregate performance across all three backbones. With Qwen2.5-7B, BCSD reaches an average success rate of 83.6 on ALFWorld, outperforming the best baseline by 3.1 points. It also achieves the highest WebShop success rate of 78.1, while its task score is only 0.2 points below the best result. The improvements are more pronounced with Qwen2.5-3B, where BCSD surpasses the strongest baselines by 5.4 points on ALFWorld and by 4.4 and 9.4 points in WebShop score and success rate, respectively. BCSD also obtains the best or tied-best results on several ALFWorld task types, indicating that its overall gains are not driven by a single category.
On Qwen3-1.7B, BCSD achieves the highest ALFWorld average success rate of 46.9, outperforming the best baseline by 3.9 points. It also obtains the highest WebShop score of 76.8, while its success rate of 50.8 is only 0.8 points below the best result. Compared with Skill\_GRPO$^*$, BCSD improves the ALFWorld average, WebShop score, and WebShop success rate by 3.9, 1.7, and 2.4 points, respectively. Since Skill\_GRPO$^*$ uses the same external SkillBank during training and evaluation, these consistent improvements suggest that BCSD benefits from internalizing skill-utilization experience through Meta-Skill-guided self-distillation, thereby applying external skills more effectively.
\subsection{Ablation Studies}

\begin{table}[t]
    \centering
    \small
    \setlength{\tabcolsep}{12pt}
    \renewcommand{\arraystretch}{1.1}
    \begin{tabular}{lcc}
        \toprule
        \textbf{Initial coefficient}
        & \textbf{\textsc{ALFWorld}}
        & \textbf{\textsc{WebShop}} \\
        \midrule
        $\lambda_0=0.5$          & 74.2 & 64.1 \\
        $\lambda_0=0.3$          & 78.1 & 61.7 \\
        \rowcolor{methodgray}
        $\lambda_0=0.1$ (Ours)   & \textbf{82.0} & \textbf{66.4} \\
        $\lambda_0=0.05$          & 77.3 & 62.5 \\
        \bottomrule
    \end{tabular}
    \caption{Effect of the initial rescaling coefficient $\lambda_0$ with the Qwen2.5-3B-Instruct. We report success rates (\%).}
    \label{tab:lambda_ablation}
\end{table}

\begin{figure}[t]
    \centering
    \includegraphics[width=0.9\columnwidth]{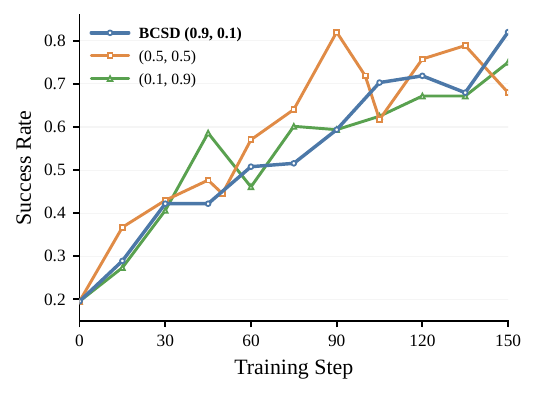}
    \caption{Effect of different weight settings on ALFWorld with Qwen2.5-3B-Instruct.}
    \label{fig:weight_ablation}
\end{figure}

\subsubsection{Component Ablation}

Table~\ref{tab:ablation} examines the contribution of each BCSD component using Qwen2.5-3B-Instruct. Removing the Meta-Skill view causes the largest degradation among the two context-view ablations, reducing performance from 82.0 to 71.1 on ALFWorld and from 66.4 to 50.0 on WebShop. This confirms the importance of policy-dependent, high-level guidance on how external skills should be applied. Removing the Pruned-Skill view also lowers performance to 75.0 and 57.0, respectively. 
This indicates that merely adding privileged information and relying on a teacher signal from a single context direction is not fully reliable, consistent with recent findings that OPD supervision quality varies across on-policy samples \citep{scope}. It also demonstrates the benefit of integrating signals from both directions.

Keeping $\lambda_n$ constant reduces performance to 76.6 on ALFWorld and 52.3 on WebShop. Early in training, a nonzero $\lambda_n$ provides dense contextual guidance when the policy is weak and outcome rewards offer limited credit assignment. Retaining the same rescaling strength throughout training, however, keeps the policy dependent on privileged-context signals even after its behavior improves, potentially interfering with later reward-driven refinement. Decaying $\lambda_n$ preserves the early benefit of self-distillation while gradually returning optimization to the environment reward.

Disabling Meta-Skill updates decreases performance to 78.1 on ALFWorld and 50.0 on WebShop, with a particularly large reduction on WebShop. A fixed Meta-Skill reflects the capabilities and failure patterns of the early policy and may become stale as its behavior evolves. Regular refreshes keep the high-level guidance aligned with the policy's current skill-utilization patterns and thus help.

\subsubsection{Hyperparameter Sensitivity}

We first study the initial rescaling coefficient $\lambda_0$ in Table~\ref{tab:lambda_ablation}. Setting $\lambda_0=0.1$ achieves the best performance on both ALFWorld and WebShop, reaching success rates of 82.0 and 66.4, respectively. Increasing it to $0.3$ or $0.5$, or decreasing it to $0.05$, consistently reduces performance. Unlike RLSD, whose privileged context provides the final ground-truth answer rather than a complete reasoning trace, BCSD uses an information-rich Meta-Skill that can affect a broad range of response tokens. A large $\lambda_0$ may therefore overemphasize context-dependent teacher signals, whereas a small value may underuse the dense guidance available early in training. These results support using a moderate initial coefficient of $0.1$ with gradual decay.

We further vary the weights assigned to the augmented and pruned gaps, denoted by $(\alpha,1-\alpha)$, in Figure~\ref{fig:weight_ablation}. The balanced $(0.5,0.5)$ setting exhibits noticeable fluctuations and fails to settle into a stable trend, suggesting that conflicting signals can exert comparable influence when the two views receive equal weights. The $(0.1,0.9)$ setting improves more slowly in the later stage, indicating that emphasizing the Pruned-Skill view may underutilize the high-level guidance provided by the Meta-Skill. In comparison, our $(0.9,0.1)$ configuration sustains stronger late-stage improvement and achieves the highest final success rate. These results support using the Meta-Skill view as the primary signal while retaining the Pruned-Skill view as a complementary reference.

\subsection{Further Analysis}

\subsubsection{Training Dynamics}

\begin{figure}[t]
    \centering
    \includegraphics[width=0.9\columnwidth]{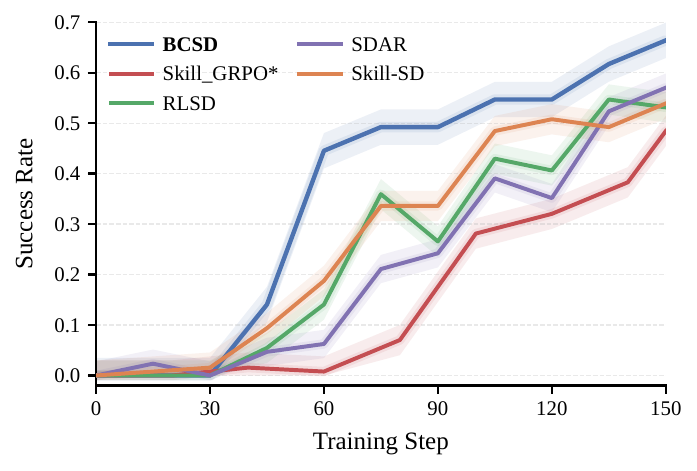}
    \caption{WebShop Success Rate on different methods with Qwen2.5-3B-Instruct.}
    \label{fig:webshop_sr}
\end{figure}

Figure~\ref{fig:webshop_sr} compares the WebShop learning curves of BCSD and the baseline methods. All methods make limited progress during the first 30 training steps. BCSD then improves substantially faster, reaching approximately $45\%$ success rate at step 60, while the other methods remain below $20\%$. After this early improvement, BCSD maintains a stable trajectory without the noticeable regressions observed in several baselines and preserves a clear lead throughout training. It ultimately reaches approximately $66\%$ success rate, outperforming the strongest baseline by around nine percentage points.

This learning pattern highlights the complementary roles of self-distillation and reward optimization in BCSD. Early in training, sparse outcome rewards provide limited token-level credit assignment, so bidirectional context signals rescale token updates while preserving the direction determined by trajectory rewards. As $\lambda_n$ decays, these signals gradually weaken and environment-derived advantages increasingly dominate. The continued improvement suggests that self-distillation serves as an effective early-stage scaffold before stable on-policy optimization takes over.

\begin{figure}[t]
    \centering
    \includegraphics[width=0.9\columnwidth]{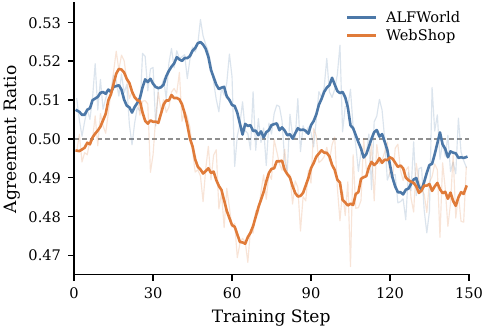}
    \caption{Token-level sign agreement ratio of two teacher of Qwen2.5-3B-Instruct on ALFWorld and WebShop.}
    \label{fig:agree_ratio}
\end{figure}

\subsubsection{Token-Level Cross-View Agreement}

We analyze whether the augmented and pruned context views provide consistent token-level guidance. At training step $k$, we define their directional agreement ratio as:
\begin{equation}
    \mathrm{Agree}_n
    =
    \frac{1}{N_n}
    \sum_{i,t}
    \mathbb{I}
    \left[
        \operatorname{sign}
        \left(\Delta_{i,t}^{\mathrm{aug}}\right)
        =
        \operatorname{sign}
        \left(\Delta_{i,t}^{\mathrm{pru}}\right)
    \right],
\end{equation}
where $N_n$ is the number of valid response tokens. As shown in Figure~\ref{fig:agree_ratio}, the agreement ratio remains close to $50\%$ throughout training on both ALFWorld and WebShop, generally fluctuating between $47\%$ and $53\%$. This indicates that the two context views provide substantially different rather than redundant token-level guidance.

For tokens on which the two views agree, their convex combination preserves the shared direction and remains bounded by the two individual gaps, avoiding additional amplification during signal fusion. For disagreeing tokens, the oppositely signed gaps partially cancel each other. Since the Pruned-Skill view receives a smaller weight, it primarily moderates rather than overrides the Meta-Skill signal, yielding more conservative guidance on controversial tokens.

\begin{figure}[t]
    \centering
    \includegraphics[width=0.9\columnwidth]{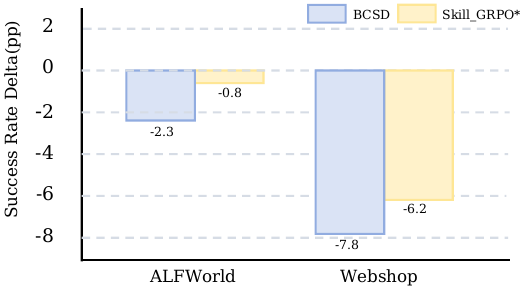}
    \caption{Success Rate delta under the General-Skill-only setting relative to the Full-Skill setting on two benchmarks.}
    \label{fig:delta}
\end{figure}

\subsubsection{Task-Skill Utilization Analysis}

Motivated by recent evaluations of whether agents effectively exploit available skills \citep{skillsbench,howwell}, we explicitly investigate whether BCSD improves the utilization of external task-specific skills rather than memorizing task-completion knowledge in its parameters.
Both BCSD and Skill\_GRPO$^*$ are trained with the Full-Skill context $C_0=(Q,G,S_c)$. After training, we remove $S_c$ and evaluate the resulting policies using only the General Skill, without further optimization. Figure~\ref{fig:delta} reports the success-rate delta between the General-Skill-only and Full-Skill settings. A larger decrease indicates that the policy's behavior depends more strongly on the task-specific guidance provided by $S_c$.

Although BCSD remains stronger in absolute performance after $S_c$ is removed, its success rate decreases more than that of Skill\_GRPO$^*$ on both benchmarks. Specifically, BCSD drops by 2.3 points on ALFWorld and 7.8 points on WebShop, compared with 0.8 and 6.2 points for Skill\_GRPO$^*$. This greater sensitivity to removing $S_c$ more clearly indicates that BCSD derives more of its gains from using the provided Task Skill without memorizing the task. If the task-completion knowledge were encoded into the model parameters, removing $S_c$ would have a smaller effect. Instead, BCSD internalizes how to use skills while keeping task-specific knowledge external and editable during inference.

\section{Conclusion}
We presented BCSD, a bidirectional context self-distillation framework for reinforcement learning of skill-based LLM agents. BCSD retains explicit skills during both training and inference and improves their utilization through two complementary context views. An augmented view provides policy-dependent Meta-Skill guidance, while a pruned view offers a focused reference for task-specific skill usage. Their token-level gaps jointly rescale the GRPO advantage without reversing the reward-determined update direction. Experiments on ALFWorld and WebShop across model scales demonstrate the effectiveness of BCSD over skill-based RL and self-distillation baselines. Ablation studies validate the contributions of both context views, Meta-Skill refresh, and coefficient decay, while further analyses show improved training dynamics and greater reliance on external skills than on memorized task knowledge. These results highlight bidirectional context self-distillation as an effective approach for training LLM agents to better utilize external skills.

\bibliography{aaai2027}


\end{document}